%% file: main.tex
\documentclass[sigconf]{acmart}
\setcopyright{none}
\acmConference[The Web Conference]{The Web Conference 2027}{2027}{}
\acmYear{2027}
\acmDOI{}
\acmISBN{}
\usepackage{graphicx}
\usepackage{booktabs,multirow,amsmath,amssymb}
\usepackage{rotating}
\usepackage{placeins}
\usepackage{algorithm}
\usepackage{algorithmic}
\newcommand{\by}{\mathbf{y}}
\newcommand{\bx}{\mathbf{x}}

\newcommand{\bw}{\mathbf{w}}
\newcommand{\bs}{\mathbf{s}}

\newcommand{\R}{\mathbb{R}}
\newcommand{\cG}{\mathcal{G}}
\newcommand{\cP}{\mathcal{P}}

\title{Compose the Horizon: Parameter-Shared Blockwise Forecasting under a Sequential-Compute Trade-off}
\author{Zeyan Li}
\affiliation{%
  \institution{Shanghai Jiao Tong University}%
  \city{Shanghai}%
  \country{China}%
}

\author{Weiheng Li}
\affiliation{%
  \institution{Jinan University}%
  \city{Guangzhou}%
  \country{China}%
}

\author{Shengda Zhuo}
\affiliation{%
  \institution{Jinan University}%
  \city{Guangzhou}%
  \country{China}%
}

\author{Libing Chen}
\affiliation{%
  \institution{The University of Chicago}%
  \city{Chicago}%
  \state{IL}%
  \country{USA}%
}

\author{Min Chen}
\affiliation{%
  \institution{South China University of Technology}%
  \city{Guangzhou}%
  \country{China}%
}

\author{Yin Tang*}
\affiliation{%
  \institution{Jinan University}%
  \city{Guangzhou}%
  \country{China}%
}

\begin{document}

\begin{abstract}
Most long-horizon forecasters emit the complete future in one pass. We ask whether a compact operator can instead be reused to emit contiguous future blocks, trading stored parameters for sequential computation. Recursive Functional Decomposition (RFD) factorizes the horizon mapping into three parameter-shared block predictions. The blocks occupy disjoint future intervals and are concatenated, not added as trend, seasonal, or residual components, and each emitted block joins the rolling context that conditions the next application. In controlled three-seed experiments, removing forward feedback hurts all four evaluated datasets while stopping gradients through it changes little, so the useful signal is the evolving forecast state itself; reapplying the shared operator helps consistently, but by less than widening it. The model uses 0.08--0.46M parameters, yet its sequential channel-wise execution is slower and more memory-intensive than an iterative baseline at high channel count. In a frozen long-context evaluation against ten strong baseline families, RFD records lower test error in 23 of 24 cells; the comparison is cross-protocol, with RFD using 512/720-step context and the baselines using 96.
\end{abstract}


\maketitle

\section{Introduction}
\label{sec:introduction}

Forecasting a distant future from historical observations is a core problem in energy, traffic, climate science, and finance. Classical methods often model trend and seasonality explicitly~\cite{hyndman2008,stl,prophet}, while many neural forecasters learn one direct mapping from a history window to the complete horizon~\cite{vaswani2017attention}. Recent long-horizon systems refine that mapping along several lines: efficient-attention Transformers over time tokens~\cite{informer,logtrans,pyraformer}, decomposition into trend and seasonal components~\cite{autoformer,fedformer,nbeats}, patch-based and channel-aware encodings~\cite{patchtst,itransformer}, and compact MLP or spectral models that keep the one-pass structure at a fraction of the parameters~\cite{dlinear,sparsetsf,fits,timemixer,wpmixer}. Across these families, accuracy and expressiveness are pursued by widening attention, feed-forward, or channel-mixing modules, so representational capacity and stored size rise together. The whole horizon mapping is stored once and applied once, no computation is reused across horizon regions, and a model expressive enough for a long horizon must also be large enough to store it. This creates a concrete systems question: can repeated application of one small operator provide a useful alternative point on the parameter--computation frontier?

\begin{figure}[t]
  \centering
  \includegraphics[width=\columnwidth]{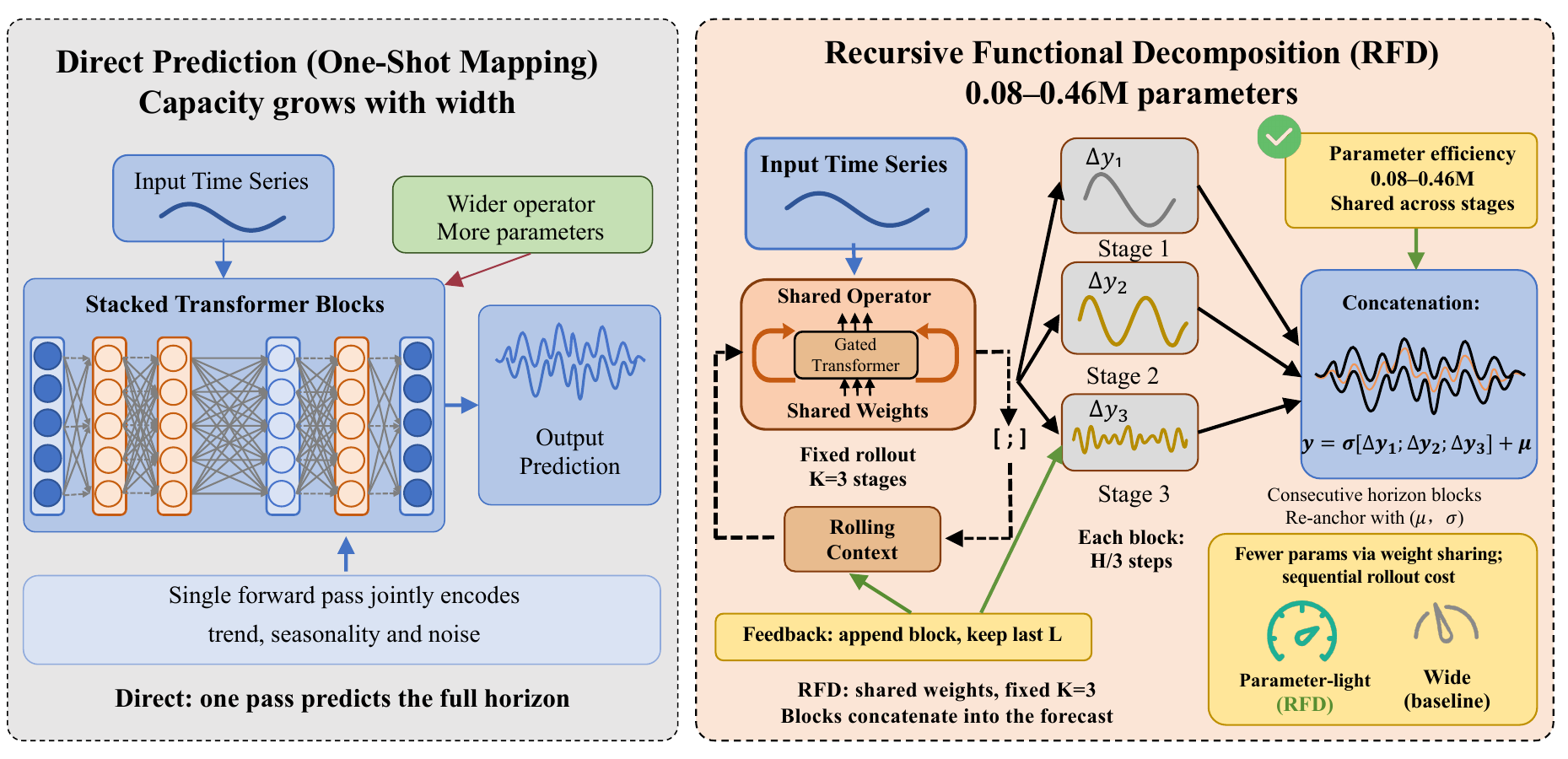}
  \caption{One-shot widening versus parameter-shared blockwise rollout. Direct prediction grows capacity through a wider single pass; RFD reuses one compact gated operator over $K{=}3$ stages, appending each emitted block to the rolling context and concatenating the blocks into the forecast.}
  \Description{Two side-by-side panels. Left: direct prediction maps the history through stacked wide Transformer blocks to the full horizon in one pass, with capacity growing through width. Right: RFD applies a small shared gated-Transformer operator three times; each application emits an H/3-step block, each block is appended to a rolling context that conditions the next stage, and the blocks are concatenated and re-anchored into the forecast within 0.08--0.46M shared parameters.}
  \label{fig:intro_motivation}
\end{figure}

Increasing model width raises storage cost. Stored parameters occupy memory at inference time, dominate the payload when a model is distributed or updated, and often scale with the horizon the model must express. At $H{=}720$, the head of a direct model must emit 720 outputs in a single pass, and its storage grows accordingly. Reusing a compact operator avoids that storage but pays in sequential compute, because each application reads the previous one and the calls cannot run in parallel. Neither axis is universally preferable; a deployment under a storage or distribution constraint benefits from the compact-operator style, while a latency-critical service favors one wide pass. The empirical question is where the accuracy--parameter--latency trade-offs of the two styles differ on standard benchmarks.

Depth--width separation results motivate testing repeated computation~\cite{cybenko1989,depth_vs_width,transformer_universal}, but they do not establish that depth is preferable for forecasting. We study a concrete construction. \textbf{Recursive Functional Decomposition (RFD)} builds the horizon one block at a time (Figure~\ref{fig:intro_motivation}): a compact update operator, reused under shared parameters, emits the next contiguous horizon block at each application, and a rolling window conditions the next application. Here ``decomposition'' means a factorization of the horizon mapping into blockwise mappings, not a trend--seasonality--residual decomposition. Unlike pointwise autoregression, each call emits a contiguous block, earlier blocks are never revised, and the number of sequential steps is fixed in advance, so the forecast is built as an evolving output-space state. The block structure also cuts the number of sequential decisions from $H$ to $K$, which shortens the autoregressive chain along which errors can compound.

We evaluate RFD on six standard datasets under univariate and multivariate protocols at horizons 96 to 720. At matched look-back length the construction is best or tied on ETTh2 and ETTm2 in both protocols and on ETTm1 in the multivariate protocol, and remains competitive on Weather and univariate Traffic. With its frozen long-context recipes it achieves lower test error in 23 of 24 dataset--horizon cells and a median reduction of $8.8\%$ against the strongest of ten baseline families. Controlled three-seed experiments then isolate the mechanism. Removing forward feedback, so that later stages cannot read earlier predictions, hurts every evaluated dataset, with mean penalties from $1.1\%$ to $80.2\%$, while stopping gradients through the same values changes results by at most $0.8\%$. The useful signal is the forecast state itself, not long-path credit assignment. Width remains the stronger capacity axis, and the sequential channel-wise execution costs throughput and memory at high channel counts, so the construction occupies one point on the frontier; it is not a free improvement.

\textbf{Architecture.} RFD is deliberately minimal. The history window is normalized by a frozen per-dataset rule and right-aligned into non-overlapping 48-step patches; one gated pre-normalization Transformer block processes each channel, and a shared head emits $H/K$ future steps. With $K{=}3$, the three blocks tile the horizon. After each stage the emitted block is appended to the context, truncated to the last $L$ steps, and visible to the next stage. The same operator and head are reused, so there is no per-stage operator replication; because the head emits $H/K$ values, the parameter count decreases as $K$ grows instead of staying constant. Operator weights are shared across channels, but optional learned input offsets have shape $L\times C$ and are channel-specific. There is no cross-variable mixer, confidence head, or learned stopping rule. Our contributions are:
\begin{itemize}
  \item A standalone parameter-shared block-rollout forecaster with right-aligned feedback, identity-gated nonlinear updates, and prefix supervision.
  \item Three-seed controls that separate coupled horizon factorization from fixed-factorization shared computation and isolate feedback information, gradient flow, objective terms, and operator choice.
  \item A joint parameter--compute account covering RFD's 0.08--0.46M range, sequential throughput, and high-channel memory.
\end{itemize}

The rest of this paper is organized as follows. Section~\ref{sec:related_work} reviews forecasting work by how it represents the horizon and spends capacity. Section~\ref{sec:approach} presents the construction, the training objective, and the parameter and complexity account. Section~\ref{sec:experiments} reports the benchmark comparison, the controlled three-seed analyses, and the efficiency profile. Section~\ref{sec:conclusion} concludes.

\section{Related Work}
\label{sec:related_work}

We organize related work by how a forecaster represents the horizon and spends capacity: widening a single direct pass, imposing additive structure, or generating the horizon autoregressively with reused computation. These distinctions matter because parameter count, accessible context, and sequential cost can change independently.

\subsection{Direct Predictors and the Width Axis}

Direct prediction learns a single mapping $f_\theta: \R^{L\times C}\!\to\!\R^{H\times C}$ in one forward pass. Early recurrent predictors captured long- and short-range dependencies, while probabilistic systems such as DeepAR separated autoregressive likelihood modeling from the scale of the model backbone~\cite{lstnet,deepar}. Transformer forecasters largely retain a fixed history-to-horizon graph. Informer, Autoformer, FEDformer, Pyraformer, and LogTrans make attention affordable through sparsity, autocorrelation, frequency-domain operations, pyramids, or locality~\cite{informer,autoformer,fedformer,pyraformer,logtrans}. PatchTST and iTransformer change what constitutes a token, while Crossformer, TimesNet, and MICN introduce cross-variable or multiscale structure~\cite{patchtst,itransformer,crossformer,timesnet,micn}. Recent mixing and MLP-style models, including WPMixer, LightTS, TimeMixer, TimeXer, and TQNet, provide strong accuracy with simpler computation~\cite{wpmixer,lightts,timemixer,timexer,tqnet}. These directions are surveyed in~\cite{wen2022transformers}.

These methods establish demanding accuracy baselines and can also benefit from longer histories. RFD does not assume that a direct predictor lacks temporal expressivity. It changes the allocation of capacity, evaluating one compact block predictor repeatedly instead of storing all transformations in one fixed pass. Consequently, comparisons must distinguish architecture from input information. We report matched-look-back results for that purpose and treat the longer-context evaluation as an end-to-end system comparison.

\subsection{Additive Structure and Fixed Bases}

An older tradition builds forecasts from explicitly additive components. Classical ARIMA, exponential smoothing, STL, and Prophet separate level, trend, seasonal, and residual effects before producing a point or interval forecast~\cite{hyndman2008,stl,prophet}. N-BEATS and N-HiTS learn stacks of backcast and forecast bases whose outputs add across blocks~\cite{nbeats,nhits}; DLinear and FITS impose low-complexity temporal or frequency projections~\cite{dlinear,fits}; and ETSformer incorporates exponential-smoothing structure~\cite{etsformer}. Related efficiency-oriented designs include SCINet, TiDE, multiscale TimeMixer, reversible normalization, and SparseTSF~\cite{scinet,tide,timemixer,revin,sparsetsf}. These methods show that useful forecasting bias need not come from a large generic Transformer.

RFD is structurally different from additive decomposition. Its stage outputs occupy disjoint future intervals and are concatenated; a later stage cannot revise or add to an earlier interval. The word ``decomposition'' refers only to factorizing the horizon mapping into block mappings.

\subsection{Autoregressive Generation and Iterative Computation}

Autoregressive forecasting already reuses computation across forecast time. DeepAR applies a recurrent transition while sampling or scoring successive future values~\cite{deepar}; WaveNet and temporal convolutional networks use causal convolutions to propagate a growing receptive field~\cite{wavenet,tcn}; and decoder-only foundation models such as TimesFM scale sequential prediction through large-scale pretraining~\cite{timesfm}. Diffusion forecasters also generate through multiple conditional steps, although those steps index denoising time, not disjoint forecast intervals~\cite{timegrad,csdi,ddpm}. Recursion itself is not our novelty; our question is narrower: can blockwise autoregression serve as a parameter-efficient capacity allocation for a small supervised forecaster? Each RFD call emits $H/K$ contiguous values, the same operator and head are reused, and the emitted values enter the raw rolling context for the next call. This lies between pointwise autoregression and direct multi-horizon prediction. It reduces the number of sequential decisions relative to pointwise generation while preserving an evolving output-space state. We also test whether this behavior is tied to one backbone by comparing linear, MLP, ungated Transformer, and gated Transformer operators under the same rollout.

A separate thread substitutes computation depth for parameter width. Deep Equilibrium Models seek a fixed point of a shared transformation~\cite{deq}; Neural ODEs parameterize continuous-depth dynamics~\cite{neural_ode}; adaptive computation time varies the number of recurrent updates~\cite{act}. These methods demonstrate that repeated application can enlarge effective computation without a proportional parameter increase, but their iterations usually refine a latent state or distribution instead of appending immutable output intervals. ReNF is the closest forecasting-specific comparison in this category. It explicitly studies an iterative neural long-term forecaster and enlarges its per-stage computation or parameterization~\cite{lu2025renf}. RFD instead fixes one block-emitting operator and reuses it at every horizon stage. Each application produces an explicit output-space block, the blocks concatenate into the forecast, and the number of applications is fixed in advance. This distinction yields fewer stored parameters but does not imply lower runtime or activation memory; we therefore report parameter count, throughput, and memory separately. Controlled feedback experiments further distinguish information flow from optimization depth, showing that later stages benefit from reading earlier predictions even when gradients through those predictions are stopped.

\section{Approach: Standalone Shared-Depth Construction}
\label{sec:approach}

\begin{figure*}[t]
  \centering
  \includegraphics[width=0.85\textwidth]{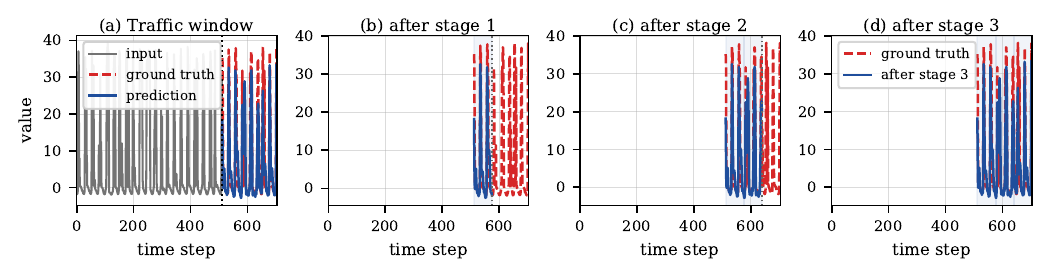}
  \caption{Traffic rollout ($L{=}512$, $H{=}192$); each stage appends one 64-step block.}
  \Description{Four plots show one Traffic example. The first compares the complete blockwise forecast with the ground truth; the next three show the forecast growing from one to three contiguous blocks as the rolling context advances.}
  \label{fig:rollout}
\end{figure*}

The goal is to predict a horizon $\by \in \R^{H \times C}$ from a history $\bx \in \R^{L \times C}$. Standard methods learn a single direct mapping $\bx \mapsto \by$; RFD instead builds the horizon in $K$ stages with one shared operator, each stage conditioned through a rolling window on the model's own predictions so far. This section presents the construction, the training objective, and the cost model.

\subsection{Overview}

\begin{figure}[t]
  \centering
  \includegraphics[width=\columnwidth]{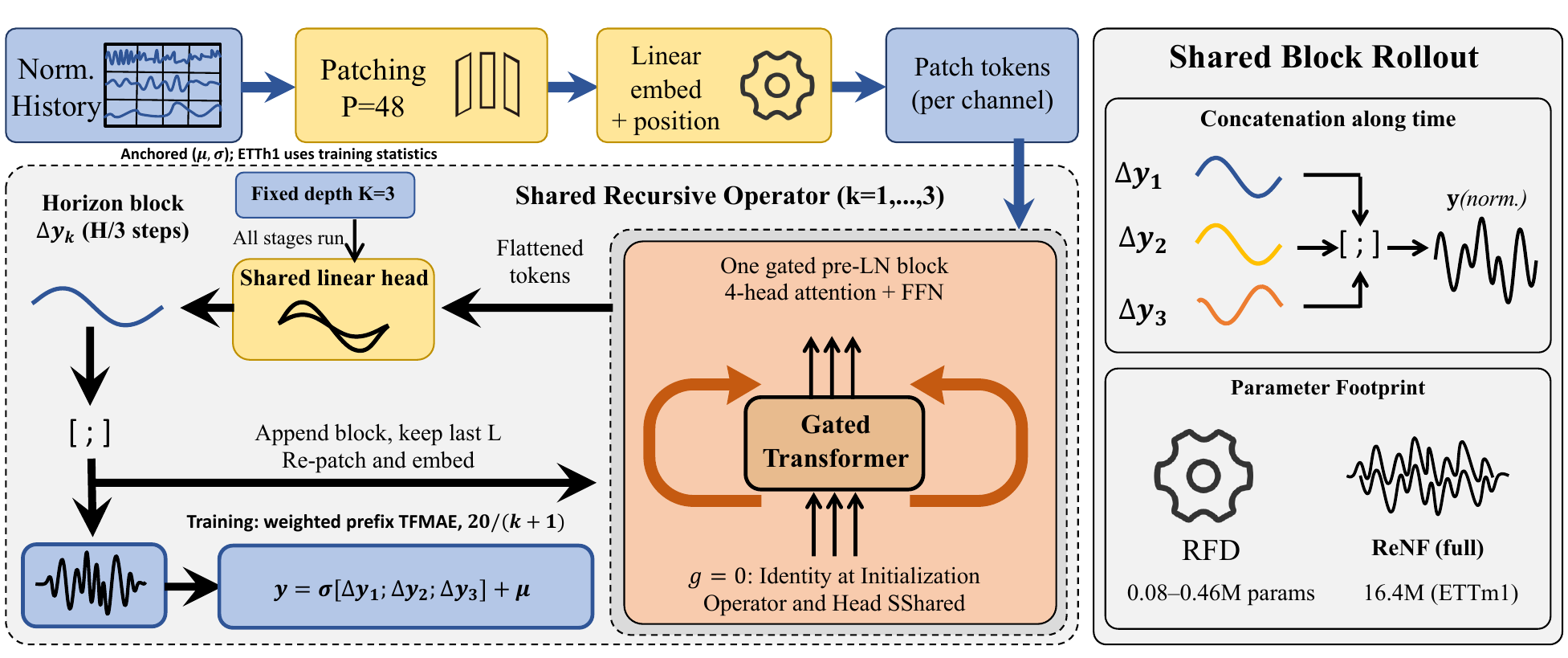}
  \caption{The RFD construction. The normalized history is right-aligned into 48-step patches; one shared recursive operator (shared linear head and one gated pre-LN block, identity at initialization) is applied $K{=}3$ times, each application emitting an $H/3$-step block that re-enters the rolling context; the blocks are concatenated and re-anchored, and training minimizes the weighted prefix TFMAE.}
  \Description{Pipeline diagram in three sections. Left: the normalized history is patched into 48-step segments and linearly embedded with positions into per-channel patch tokens. Middle: a shared recursive operator applied at stages k = 1 to 3, containing a shared linear head and one gated pre-LayerNorm Transformer block with four-head attention and a feed-forward network, initialized as the identity, emitting one horizon block of H/3 steps per stage. Right: the three blocks are concatenated along time into the normalized forecast, re-anchored with the normalization statistics, and the parameter footprint compares RFD's 0.08--0.46M parameters with ReNF's 16.4M on ETTm1.}
  \label{fig:architecture}
\end{figure}

RFD replaces a single wide predictor with a narrow predictor applied a fixed number of times (Figure~\ref{fig:architecture}). The forecast is the concatenation of the stage blocks,
\begin{equation}
\hat{\by} = [\Delta\hat{\by}_{1}; \Delta\hat{\by}_{2}; \dots; \Delta\hat{\by}_{K}],
\label{eq:overview}
\end{equation}
where each $\Delta\hat{\by}_k \in \R^{(H/K)\times C}$ is produced by the shared operator at stage $k$ from a rolling context that already contains the blocks emitted so far. Because the operator parameters are reused at every stage, increasing depth does not create a new operator per stage. In this implementation the shared head becomes smaller as $K$ grows because its output width is $H/K$, while computation grows with depth. Figure~\ref{fig:intro_motivation} illustrates the principle.

This organization factorizes a long output mapping into shorter block mappings. Rolling feedback exposes earlier predictions to later stages, and parameter sharing avoids a separate parameter set for every stage. Whether this factorization is preferable to a direct predictor is an empirical question, not a consequence of the construction alone.

\subsection{Context Encoding}
\label{sec:encoding}

The updater receives a compact patch representation of the rolling window, which reduces the token count. Each window is normalized by anchored statistics. We use the mean and standard deviation of the last $H$ observations, without affine parameters and detached from the computation graph:
\begin{equation}
\mu \;=\; \tfrac{1}{H}\!\!\sum_{i=L-H+1}^{L}\! \bx_i,
\qquad
\sigma^{2} \;=\; \tfrac{1}{H}\!\!\sum_{i=L-H+1}^{L}\! (\bx_i - \mu)^{2},
\qquad
\bw \;=\; \frac{\bx - \mu}{\sigma + \epsilon},
\label{eq:anchor}
\end{equation}
and right-aligned into non-overlapping patches of length $P{=}48$. Let $N=\lfloor L/P\rfloor$ and $r=L-NP$. The implementation discards the \emph{oldest} $r$ values and retains the most recent $NP$ values, giving ten patches at $L{=}512$ ($r{=}32$) and fifteen at $L{=}720$ ($r{=}0$). A linear embedding with per-patch position parameters maps each patch into $\R^{d}$,
\begin{equation}
\bs_i \;=\; W_e\, \bw_{r+(i-1)P+1\,:\,r+iP} \;+\; \mathbf{e}_i,
\qquad i = 1, \dots, N,
\quad N = \lfloor L/P \rfloor,
\label{eq:patch_embed}
\end{equation}
yielding one token sequence $\bs_{1:N} \in \R^{N \times d}$ per channel. Right alignment is essential for feedback: with $L{=}512,H{=}96,K{=}3$, Stage 2 encodes $[x_{65:512};\Delta\hat{\by}_1]$, so all 32 values emitted by Stage 1 enter the next computation. One dataset (ETTh1) uses global training-split standardization instead; the choice was made by the validation-only selection described in Section~\ref{sec:exp_setup}. For recipes with positional offsets enabled, an additional learned tensor $Q\in\R^{L\times C}$ is added to the rolling window before patching. $Q$ is channel-specific and is the only source of channel-count-dependent parameters; it performs no cross-channel mixing.

\subsection{Shared Recursive Updater}
\label{sec:updater}

The updater works by successive horizon extension. Stage $k$ emits the $k$th contiguous future interval; the next stage sees that block through the rolling window and emits the following interval.

Let $\cG_\theta$ be one pre-LayerNorm Transformer block with four self-attention heads and a $d\!\to\!2d\!\to\!d$ FFN, applied to the patch tokens of one channel, and let $\cP_\psi:\R^{N d}\to\R^{H/K}$ be a linear head that reads the flattened patch tokens. Channels are folded into the batch axis and use the same operator weights; the optional $Q\in\R^{L\times C}$ input offsets remain channel-specific, and there is no cross-variable mixer. Both residual branches are scaled by $\tanh$ of the same zero-initialized scalar $g$,
\begin{equation}
\begin{aligned}
\widetilde{\bs} &= \bs + \tanh(g)\,\mathrm{Attn}(\mathrm{LN}_1(\bs)),\\
\bs' &= \widetilde{\bs} + \tanh(g)\,\mathrm{FFN}(\mathrm{LN}_2(\widetilde{\bs})),
\end{aligned}
\label{eq:gated_block}
\end{equation}
so at initialization ($g{=}0$) the gated block is the identity. The surrounding patch embedding, position offsets, and prediction head are affine, so the initialized rollout is affine in normalized coordinates. Applying this block and the shared head once per stage, with every emitted block fed back into the context, defines the full rollout
\begin{equation}
\begin{aligned}
\bs^{(k)} &= \cG_\theta\bigl(\mathrm{Patch}(\bw_{k-1})\bigr),\\
\Delta\hat{\by}_k &= \cP_\psi\bigl(\mathrm{flatten}(\bs^{(k)})\bigr),\\
\bw_k &= [\bw_{k-1};\, \Delta\hat{\by}_k]_{:,\,-L:},
\end{aligned}
\qquad k = 1, \dots, K,
\label{eq:rollout}
\end{equation}
starting from the normalized history $\bw_0 = \bw$; all blocks live in normalized coordinates, and the forecast is re-anchored as $\hat{\by} = \sigma\,[\Delta\hat{\by}_1; \dots; \Delta\hat{\by}_K] + \mu$, which recovers Equation~\ref{eq:overview}. Whether positional offsets are added to the rolling window is a per-dataset choice fixed by the same validation-only selection (Section~\ref{sec:exp_setup}). Algorithm~\ref{alg:inference} summarizes inference, and training executes the identical rollout.

Because $\cG_\theta$ and $\cP_\psi$ are shared across all $K$ stages, no stage-specific operator parameters are introduced. The head output width is $H/K$, however, so changing $K$ changes the head size and makes the total parameter count non-increasing in the depth sweep. The block-emitting update is the key inductive bias: each stage only extends the forecast and cannot overwrite earlier blocks, while the shared parameters prevent independent per-stage specialization.

Figure~\ref{fig:rollout} visualizes the construction on a Traffic test window; each stage appends one block, and the shaded prefix is what later stages re-read.

\input{avg_table_s.tex}

\input{avg_table_ms.tex}

\subsection{Fixed-Depth Training}

The building block of the objective is a convex combination of pointwise error and the magnitude of the complex spectral residual,
\begin{equation}
\mathrm{TFMAE}_{\alpha}(\hat{\by}, \by) \;=\; (1-\alpha)\, \mathrm{MAE}(\hat{\by}, \by) \;+\; \alpha\, \operatorname{mean}\!\left|\mathcal{F}(\hat{\by})-\mathcal{F}(\by)\right|,
\label{eq:tfmae}
\end{equation}
where $\mathcal{F}$ is the real FFT along the time axis, $|\cdot|$ is complex magnitude, and the weight $\alpha\in[0,0.5]$ is fixed per dataset. This loss penalizes phase as well as amplitude disagreement; it is not an MAE between magnitude spectra. The training objective supervises every stage prefix, not only the final output,
\begin{equation}
\mathcal{L}_{\mathrm{roll}} \;=\; \sum_{k=1}^{K} \frac{20}{k}\; \mathrm{TFMAE}\bigl(\hat{\by}_{:kH/K},\; \by_{:kH/K}\bigr),
\label{eq:prefix_loss}
\end{equation}
so early horizon intervals receive the largest direct weight. For $K{=}3$, repeated prefix averaging makes the per-time-step coefficients of the three blocks proportional to $12.25{:}3.25{:}1$ for the pointwise term. This is a near-horizon-favoring heuristic; its contribution is isolated stepwise in Section~\ref{sec:rq3}. Optimization uses Adam with a step-decay schedule, and the selected checkpoint is the one minimizing full validation MSE. Every training example executes the full $K$ stages; there is no confidence head, learned stopping rule, or input-dependent exit, and the training code has no access to the test split.

\begin{algorithm}[t]
\caption{RFD Inference (block rollout, fixed depth $K$)}
\label{alg:inference}
\small
\begin{algorithmic}[1]
\REQUIRE History $\bx \in \R^{L \times C}$; frozen normalization rule $\mathcal{N}$ and inverse $\mathcal{N}^{-1}$
\REQUIRE Patch encoder $\textsc{Patch}$, updater $\cG_\theta$, head $\cP_\psi$, depth $K$
\ENSURE Forecast $\hat{\by} \in \R^{H \times C}$
\STATE $\bw \leftarrow \mathcal{N}(\bx)$; $\bs \leftarrow \textsc{Patch}(\bw)$ \COMMENT{anchored; global statistics for ETTh1}
\STATE $\hat{\by} \leftarrow$ empty
\FOR{$k=1$ to $K$}
    \STATE $\bs \leftarrow \cG_\theta(\bs)$ \COMMENT{gated block; identity at init}
    \STATE $\Delta\by_k \leftarrow \cP_\psi(\mathrm{flatten}(\bs))$
    \STATE $\hat{\by} \leftarrow [\hat{\by}; \Delta\by_k]$;\; $\bw \leftarrow [\bw; \Delta\by_k]_{:,-L:}$ \COMMENT{rolling feedback}
    \STATE $\bs \leftarrow \textsc{Patch}(\bw)$
\ENDFOR
\RETURN $\mathcal{N}^{-1}(\hat{\by})$
\end{algorithmic}
\end{algorithm}

A forward pass costs $\mathcal{O}(K(Nd^{2}+N^{2}d) + N d H)$ per channel: $K$ applications of the shared FFN and self-attention block plus a head whose total output width across stages equals $H$. The total parameter count is
{\small
\begin{equation}
|\Theta| = \underbrace{\Theta(d^{2})}_{\text{attention + FFN}}
+ \underbrace{N d \cdot H/K}_{\text{head}}
+ \underbrace{\mathbb{I}_{\mathrm{pe}}LC}_{\text{optional input offsets}}
+ \underbrace{\Theta(Pd+Nd)}_{\text{patch embedding/positions}}.
\label{eq:param_count}
\end{equation}
}
with no factor of $K$ multiplying the operator parameters and a head term that decreases as $K$ grows. Thus the count is non-increasing, not constant, in the stage-count sweep. The optional $LC$ term explains why Table~\ref{tab:rq4_efficiency} has more parameters at $C{=}862$ than at $C{=}7$. In absolute terms the model uses 0.08--0.19M parameters at $H{=}96$ and at most 0.46M over the entire $6{\times}4$ grid; the per-stage-widening iterative forecaster ReNF~\cite{lu2025renf} reaches 16.4M parameters on ETTm1 in our local reproduction of its official configuration, roughly $90\times$ ours on the same cell.

\section{Experiments}
\label{sec:experiments}

This section evaluates the construction of Section~\ref{sec:approach}. Development never touches the test split; the test-stage comparisons in Section~\ref{sec:benchmark} were run only after their respective configurations were frozen. Four research questions organize the evaluation:
\begin{itemize}
    \item[RQ1] How does the construction compare with standard forecasters under matched-look-back and full long-context test protocols?
    \item[RQ2] When parameters are reused, what is gained by horizon factorization and by additional shared computation relative to width?
    \item[RQ3] Which information flow, objective, and operator choices make blockwise rollout effective?
    \item[RQ4] What is the parameter, memory, and throughput profile of the construction relative to a per-stage-widening iterative forecaster, and how does it scale with the channel count?
\end{itemize}

\subsection{Experimental Setup}
\label{sec:exp_setup}

\begin{table*}[t]
    \caption{Long-context test MSE against the strongest baseline per cell from Tables~\ref{tab:baseline_avg}--\ref{tab:ms_avg}.}
    \label{tab:test_multiseed}
    \begin{center}
    \begin{small}
    \setlength{\tabcolsep}{7pt}%
    \begin{tabular}{ll llll}
    \toprule
    & & $H{=}96$ & $H{=}192$ & $H{=}336$ & $H{=}720$ \\
    \midrule
    ETTh1 & RFD & \textbf{.3737} & \textbf{.4125} & \textbf{.4472} & \textbf{.4613} \\
          & best baseline & .3749 (WPMixer) & .4209 (FEDformer) & .4666 (WPMixer) & .4968 (WPMixer) \\
    \addlinespace[2pt]
    ETTh2 & RFD & \textbf{.2815} & \textbf{.3408} & \textbf{.3635} & .4021 \\
          & best baseline & .2912 (TimeMixer) & .3729 (TQNet) & .4155 (TimeXer) & \textbf{.4020 (TimeXer)} \\
    \addlinespace[2pt]
    ETTm1 & RFD & \textbf{.2871} & \textbf{.3208} & \textbf{.3602} & \textbf{.4179} \\
          & best baseline & .3172 (WPMixer) & .3648 (WPMixer) & .3980 (WPMixer) & .4597 (WPMixer) \\
    \addlinespace[2pt]
    ETTm2 & RFD & \textbf{.1592} & \textbf{.2173} & \textbf{.2643} & \textbf{.3587} \\
          & best baseline & .1740 (WPMixer) & .2395 (WPMixer) & .2976 (TimeMixer) & .3991 (TimeMixer) \\
    \addlinespace[2pt]
    Weather & RFD & \textbf{.1871} & \textbf{.2400} & \textbf{.2954} & \textbf{.3825} \\
          & best baseline & .2129 (TimeXer) & .2756 (TimeXer) & .3319 (TimeXer) & .4224 (TimeXer) \\
    \addlinespace[2pt]
    Traffic & RFD & \textbf{.3746} & \textbf{.3901} & \textbf{.4018} & \textbf{.4328} \\
          & best baseline & .3967 (iTransformer) & .4195 (iTransformer) & .4350 (iTransformer) & .4664 (iTransformer) \\
    \bottomrule
    \end{tabular}
    \end{small}
    \par\smallskip
    \footnotesize RFD is the three-seed mean at $L\in\{512,720\}$; baselines use $L{=}96$. Bold marks the lower value.
    \end{center}
\end{table*}

Evaluation uses six benchmarks from the Time Series Library~\cite{timeserieslib}: Weather (21 climate variables), ETTh1 and ETTh2 (hourly), ETTm1 and ETTm2 (15-minute), and Traffic (862 road sensors), with horizons $H \in \{96,\allowbreak 192,\allowbreak 336,\allowbreak 720\}$. Splits and data access follow the library's official configuration, and all preprocessing and training use only training and validation data. For Weather we interpolate the 81 missing values that occur in the training period. The full RFD system uses per-dataset context lengths $L\in\{512,720\}$; the matched-look-back comparison uses $L{=}96$. Every locally trained result reported in the paper uses seeds 2021/2022/2023, selects checkpoints by full validation MSE, and reports the arithmetic mean. Sample standard deviations appear in Figures~\ref{fig:rq2_shared_depth} and~\ref{fig:rq3_controlled} and in the controlled tables of Appendix~\ref{app:controlled}. Test evaluation is a separate frozen-checkpoint pass and is never used for recipe or checkpoint selection.

\subsection{Standard Benchmark Comparison (RQ1)}
\label{sec:benchmark}

RQ1 asks whether the parameter-shared construction is competitive on standard forecasting benchmarks. We report the univariate (S) and multivariate (M) protocols at horizons 96, 192, 336, and 720, with results averaged over horizons per dataset (lower is better). The matched-look-back tables control input length at $L{=}96$; the subsequent long-context table evaluates the complete RFD system at $L\in\{512,720\}$. Baseline values come from official releases or faithful re-implementations on the common benchmark splits; protocol details can still differ across releases, so the tables state the look-back scope explicitly. Tables~\ref{tab:baseline_avg} (S) and~\ref{tab:ms_avg} (M) summarize the matched-look-back comparison; Appendix~\ref{app:full_results} provides per-horizon breakdowns.

The matched comparison establishes competitiveness and its limits. RFD is best or tied on ETTh2 and ETTm2 in both protocols and on ETTm1 in the multivariate protocol; its univariate ETTm1 average trails the best baseline by $0.001$. It is competitive on Weather and univariate Traffic. It is not uniformly dominant. ETTh1 univariate forecasting favors several direct models, and multivariate Traffic favors iTransformer by a substantial margin. The recurring strength on the ETT minute-level datasets shows that a small shared rollout can match strong direct predictors, while the failures rule out the explanation that recursion alone produces a universal advantage.

Table~\ref{tab:test_multiseed} then evaluates the frozen long-context system. Its per-cell baseline is the strongest of the ten families already present in Tables~\ref{tab:baseline_avg}--\ref{tab:ms_avg}. RFD records 23/24 \emph{numerical} wins, with a median reduction of $8.8\%$ from the selected baseline. The largest conceptual change is Traffic, where the same model family that trails at matched $L{=}96$ leads after using the longer history. This reversal is useful but not causal evidence for the architecture, because both accessible context and model differ. Accordingly, Tables~\ref{tab:baseline_avg}--\ref{tab:ms_avg} answer whether the construction is competitive with matched information, whereas Table~\ref{tab:test_multiseed} measures the end-to-end long-context system. Together they establish practical competitiveness and long-context utilization.

\subsection{Stage Factorization versus Width (RQ2)}
\label{sec:rq2}

RQ2 first compares two ways of changing capacity at $L{=}H{=}96$ on the four ETT datasets. One arm varies $K\in\{1,2,3,4,6\}$ at fixed $d_{\mathrm{ff}}{=}128$; the other varies $d_{\mathrm{ff}}\in\{16,32,64,128\}$ at fixed $K{=}3$. Changing $K$ also changes block length $H/K$, head size, feedback frequency, and prefix-loss geometry. We therefore call it the \emph{stage-factorization arm}, not an isolated depth intervention. A second control, reported in full in Appendix~\ref{app:controlled}, holds those factors fixed and varies only the number of shared operator applications inside each stage, on ETTh1, ETTm2, and Traffic at $H\in\{96,336\}$.

\begin{table}[t]
    \caption{RQ2 stage-factorization and width sweeps on the four ETT datasets (three-seed validation MSE).}
    \label{tab:rq2_depth_width}
    \centering
    \begin{small}
    \setlength{\tabcolsep}{2.0pt}%
    \resizebox{\columnwidth}{!}{%
    \begin{tabular}{lccccc@{\hspace{5pt}}cccc}
    \toprule
    & \multicolumn{5}{c}{Stage factorization ($d_{\mathrm{ff}}{=}128$)} &
    \multicolumn{4}{c}{Width ($K{=}3$)} \\
    \cmidrule(lr){2-6}\cmidrule(lr){7-10}
    Dataset & $K{=}1$ & $K{=}2$ & $K{=}3$ & $K{=}4$ & $K{=}6$ &
    $d_{\mathrm{ff}}{=}16$ & $32$ & $64$ & $128$ \\
    \midrule
    ETTm1 & .3816 & .3817 & .3814 & \textbf{.3801} & .3817 &
    .3900 & .3860 & .3832 & \textbf{.3814} \\
    ETTm2 & \textbf{.1208} & .1217 & .1216 & .1218 & .1210 &
    .1234 & .1229 & .1225 & \textbf{.1216} \\
    ETTh2 & \textbf{.2109} & .2129 & .2134 & .2146 & .2146 &
    .2147 & .2133 & \textbf{.2127} & .2134 \\
    ETTh1 & .6813 & .6734 & .6702 & .6673 & \textbf{.6664} &
    .7482 & .6962 & .6816 & \textbf{.6702} \\
    \midrule
    params (k) & 164 & 152 & 148 & 146 & 144 &
    5 & 13 & 42 & 148 \\
    \bottomrule
    \end{tabular}
    }
    \end{small}
\end{table}

\begin{figure*}[t]
  \centering
  \includegraphics[width=\textwidth]{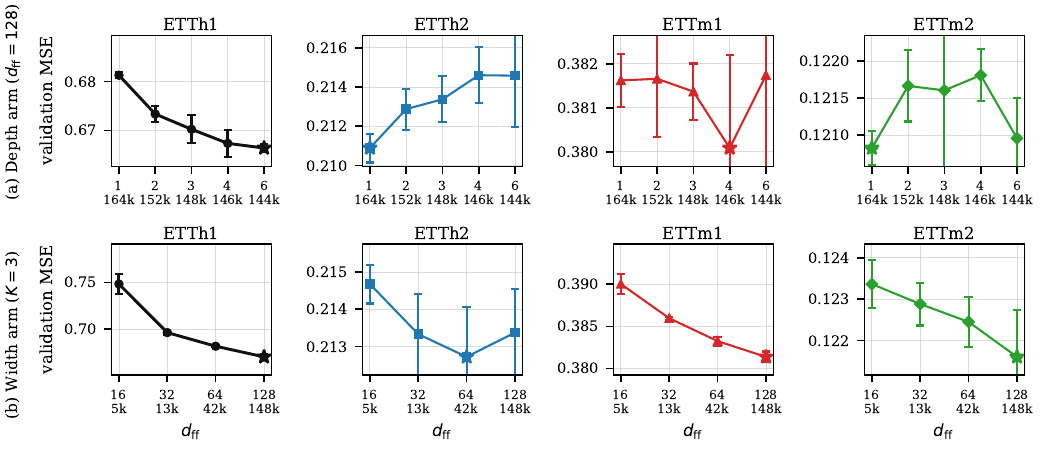}
  \caption{RQ2 sensitivity across three seeds; error bars show sample standard deviation.}
  \Description{Eight line plots in a two-by-four layout. The top row varies stage count and the bottom row varies width for ETTh1, ETTh2, ETTm1, and ETTm2; points show three-seed mean validation MSE with standard-deviation error bars.}
  \label{fig:rq2_shared_depth}
\end{figure*}

The two interventions in Table~\ref{tab:rq2_depth_width} give different pictures. In the coupled sweep, ETTh1 improves as the horizon is divided more finely while the shared head shrinks, but ETTm1 changes little and neither ETTm2 nor ETTh2 benefits. The width sweep is more regular, improving three datasets monotonically and producing a much larger change on ETTh1. In numbers, widening from $d_{\mathrm{ff}}{=}16$ to $128$ lowers ETTh1 validation MSE by $10.4\%$, whereas the entire stage-factorization sweep spans $2.2\%$ on ETTh1 and under half a percent on ETTm1. Figure~\ref{fig:rq2_shared_depth} makes this contrast visible.

The fixed-factorization control resolves the main ambiguity. Across 12 dataset--horizon--width cells, the best repeated setting improves every cell relative to one application, but only by $0.01$--$0.72\%$; four applications are not always better than two. The effect persists at $H{=}336$ on ETTh1 and Traffic, so it is not restricted to the shortest horizon, yet it remains smaller than the width effect. Repeated shared computation is therefore a measurable secondary capacity axis, while the larger gains of stage factorization can also arise from shortening each block and changing the head. The answer to RQ2 is not ``depth replaces width'': shared computation complements width and can improve the parameter--accuracy frontier without universally dominating it.

The parameter trajectories sharpen this interpretation. Increasing $K$ reduces the head from 96 outputs toward shorter blocks, so the ETTh1 gain occurs while the parameter count falls; it cannot be explained by storing a larger model. The same reduction fails to help ETTh2, however, which shows that the useful intervention is not additional abstract depth alone. It changes the prediction problem: later calls solve shorter conditional extensions from a context containing model-generated values. The fixed-factorization control removes that task change and leaves only more applications of the same transformation. Its much smaller effect indicates that most of the coupled sweep's movement comes from how the horizon is partitioned, not merely from executing more layers. This distinction also changes how the curves should be used. Several repeated-compute gains are comparable to the observed seed variation, so they motivate a capacity option, not a universal ranking. Width gives a larger and more regular accuracy response, while shared repetition changes computation without adding stage-specific parameters. A practitioner can widen when memory permits, or spend additional sequential compute when parameter storage is constrained. RFD's scientific contribution is to expose this second choice and connect it to an evolving forecast state.

\subsection{Mechanism and Design Controls (RQ3)}
\label{sec:rq3}

Table~\ref{tab:rq3_ablation} changes several mechanisms at once and is best read as a stress test. A1 opens the residual gate. A2 removes rolling feedback but keeps the weights shared. With no stage identifier, every stage receives the same input and repeats the same block. A3 simultaneously removes prefix supervision, changes MAE to MSE, and removes the spectral term. These failures identify questions, not causes, so we follow them with matched controls for feedback, gradient flow, loss composition, and operator choice.

\begin{table}[b]
    \caption{RQ3 bundled stress tests on the four ETT datasets; A0 is the full RFD model (three-seed validation MSE).}
    \label{tab:rq3_ablation}
    \begin{center}
    \begin{small}
    \setlength{\tabcolsep}{4pt}%
    \begin{tabular}{l c cc cc cc}
    \toprule
    & A0 & \multicolumn{2}{c}{A1 (gate open)} & \multicolumn{2}{c}{A2 (no feedback)} & \multicolumn{2}{c}{A3 (plain MSE)} \\
    \cmidrule(lr){2-2}\cmidrule(lr){3-4}\cmidrule(lr){5-6}\cmidrule(lr){7-8}
    Dataset & MSE & MSE & $\Delta\%$ & MSE & $\Delta\%$ & MSE & $\Delta\%$ \\
    \midrule
    ETTm1 & .3812 & .3810 & 0.0 & .9125 & $+139.4$ & .3918 & $+2.8$ \\
    ETTm2 & .1216 & .1203 & $-1.1$ & .1552 & $+27.6$ & .1269 & $+4.3$ \\
    ETTh2 & .2134 & .2137 & $+0.1$ & .2523 & $+18.2$ & .2144 & $+0.5$ \\
    ETTh1 & .6699 & .6669 & $-0.4$ & 1.2057 & $+80.0$ & .6631 & $-1.0$ \\
    \bottomrule
    \end{tabular}
    \end{small}
    \end{center}
\end{table}

\begin{figure*}[t]
  \centering
  \includegraphics[width=\textwidth]{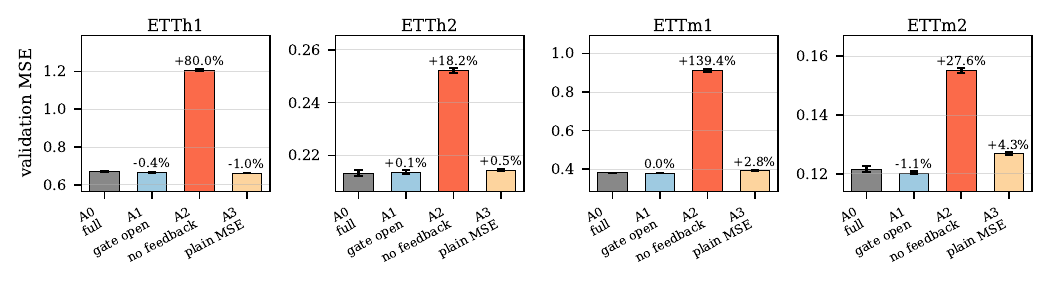}
  \caption{RQ3 bundled stress tests; error bars show sample standard deviation.}
  \Description{Four bar charts for ETTh1, ETTh2, ETTm1, and ETTm2. Each compares the full model, open-gate initialization, unconditioned no-feedback, and final MSE using three-seed mean validation MSE and standard-deviation error bars.}
  \label{fig:rq3_controlled}
\end{figure*}

Figure~\ref{fig:rq3_controlled} shows the stress-test comparison. Opening the gate is nearly neutral at convergence, the degenerate A2 fails sharply, and the bundled objective change has no consistent sign. The decisive feedback comparison instead equips every stage with an explicit identifier for which horizon block it emits, and changes only whether earlier predictions are present. Removing those predictions hurts all 12 paired-seed runs (four ETT datasets $\times$ three seeds), with mean penalties from $1.1\%$ on ETTm2 to $80.2\%$ on ETTh1. Detaching the same forward values changes mean MSE by at most $0.8\%$. Later stages therefore use information in the running forecast; long-path gradient flow through that forecast is not the main explanation.

The objective and operator controls bound how specialized the realization must be. Stepwise loss decomposition on the four ETT datasets shows that TFMAE is best on both 15-minute datasets at the tested horizons, with its clearest spectral contribution at $H{=}720$, but it is not best on the hourly datasets where the spectral weight is zero. Likewise, under the same rollout on ETTh1, ETTm2, and Traffic, a Transformer variant leads four of six dataset--horizon cells, while a linear or MLP operator leads both ETTm2 cells. The reusable principle is thus the evolving blockwise forecast state, not a universal claim for the gate, Transformer block, or spectral loss. In practice this makes the objective and the operator the two components worth re-selecting when the temporal regime changes, while the rollout itself can stay fixed. Appendix~\ref{app:controlled} reports every controlled mean and sample standard deviation.

These controls explain the A2 gap: its stages lacked both feedback and a stage identifier, so failure combined missing forecast state with an inability to distinguish horizon blocks. Once stage identity is restored, removing feedback still hurts every paired run, but the magnitude becomes dataset dependent: it is decisive on the hourly series and modest on the minute-level series. The forward state is useful information whose value depends on the temporal regime; it is not a universally dominant autoregressive effect.

The stop-gradient result further localizes the mechanism. Training later blocks through earlier predictions is not required to preserve most of the benefit; what matters is that the predictions are present in the forward computation. This does not remove the sequential inference cost, but it separates state construction from long-path credit assignment. Together with the operator study, it suggests a robust design rule: preserve the rolling output-space state and shared block interface, while treating gating, backbone, and objective as replaceable implementation choices. That rule explains both the strong feedback effect and the absence of one universally best operator or loss.

\subsection{Efficiency Profile (RQ4)}
\label{sec:rq4}

RQ4 measures the compute footprint and its scaling with the channel count $C$: $L{=}H{=}96$, batch 32, 20 timed iterations on a single NVIDIA H20. We compare our model across widths against ReNF in its lightweight ``beta'' configuration, sized to keep its parameter count comparable to ours (Appendix~\ref{app:implementation}), at $C{=}7$ (the ETT regime) and $C{=}862$ (the Traffic regime), reporting parameter count, peak device memory, and training throughput. Table~\ref{tab:rq4_efficiency} summarizes the profile.

\begin{table}[t]
    \caption{RQ4 parameter and compute profile ($L{=}H{=}96$).}
    \label{tab:rq4_efficiency}
    \begin{center}
    \begin{small}
    \setlength{\tabcolsep}{6pt}%
    \begin{tabular}{ll rrr}
    \toprule
    $C$ & Model & params (k) & mem (MiB) & thr (smp/s) \\
    \midrule
    \multirow{5}{*}{$7$}
      & RFD ($d_{\mathrm{ff}}{=}16$)  & 4.8   & 81.3 & 4788 \\
      & RFD ($d_{\mathrm{ff}}{=}32$)  & 12.9  & 82.6 & 4725 \\
      & RFD ($d_{\mathrm{ff}}{=}64$)  & 41.5  & 85.4 & 4827 \\
      & RFD ($d_{\mathrm{ff}}{=}128$) & 147.9 & 91.7 & 4798 \\
      & ReNF (beta)                    & 1101.8 & 86.4 & 10310 \\
    \midrule
    \multirow{5}{*}{$862$}
      & RFD ($d_{\mathrm{ff}}{=}16$)  & 86.8  & 2067.6 & 548 \\
      & RFD ($d_{\mathrm{ff}}{=}32$)  & 95.0  & 2217.5 & 537 \\
      & RFD ($d_{\mathrm{ff}}{=}64$)  & 123.6 & 2523.8 & 510 \\
      & RFD ($d_{\mathrm{ff}}{=}128$) & 230.0 & 3104.3 & 458 \\
      & ReNF (beta)                    & 1183.9 & 829.4 & 3245 \\
    \bottomrule
    \end{tabular}
    \end{small}
    \end{center}
\end{table}

Parameters and compute tell opposite stories. At the headline width the model stores roughly one eighth as many parameters as ReNF (beta) at $C{=}7$ and one fifth as many at $C{=}862$. The increase across channel counts is fully explained by the optional $L\times C$ input offsets; the operator remains shared, but the complete model is not channel-count independent. Runtime reverses the comparison. Folding channels into the batch axis materializes many channel-wise activations, and every forecast requires sequential stage calls; on Traffic this yields several times the peak memory and substantially lower throughput than ReNF (beta). RFD is therefore attractive when model storage or parameterized capacity is the constraint, not when training throughput or activation memory is the primary constraint. Parameter efficiency is not compute efficiency.

\subsection{Synthesis across the Four Questions}

The experiments form one causal chain. RQ1 shows the construction is competitive with matched information and exploits longer histories; RQ2 shows that neither stage count nor repeated computation explains the whole effect and that width remains important; RQ3 identifies the strongest internal link in information flow, as later stages need the forecast state produced by earlier stages, with optimizer path, loss, and operator as contingent choices; RQ4 prices the behavior in sequential execution and high-channel activation cost. A long horizon can therefore be constructed as an evolving output-space state with one reusable predictor, a parameter--computation trade-off with an isolated information-flow mechanism.


\section{Conclusion}
\label{sec:conclusion}

We presented RFD, a parameter-shared blockwise forecaster that composes the horizon from three contiguous blocks emitted by one compact operator, with right-aligned patching so that every emitted block re-enters the next stage. The forecast itself is the evolving state of the construction.

The controls identify what makes this work. Later stages use the forecast state. Removing forward feedback hurts every paired run once stage identity is fixed, while stopping gradients through those values changes little. Reapplying the shared operator adds a real but secondary capacity axis (at most $0.72\%$ in controlled cells), with width remaining the stronger axis. The principle is portable: linear, MLP, and Transformer operators all work inside the same rollout, with the gated Transformer a strong default but not a requirement.

At $0.08$--$0.46$M parameters, RFD matches strong direct predictors at matched look-back and achieves lower test error in 23 of 24 cells with the frozen long-context system. Parameter-shared blockwise forecasting is therefore a practical alternative to one-shot widening. It trades stored capacity for sequential compute, and the trade favors deployments in which stored capacity, not wall-clock latency, is the binding constraint. A fully matched long-context retraining of the strongest baselines is the natural next step for isolating the architectural contribution. Generative AI assisted with language editing and formatting; the authors verified and take responsibility for all content.

\bibliographystyle{ACM-Reference-Format}
\bibliography{paper}

\clearpage
\appendix
\setcounter{topnumber}{4}
\setcounter{dbltopnumber}{3}
\setcounter{totalnumber}{6}
\renewcommand{\topfraction}{0.95}
\renewcommand{\dbltopfraction}{0.95}
\renewcommand{\textfraction}{0.05}
\renewcommand{\floatpagefraction}{0.85}
\renewcommand{\dblfloatpagefraction}{0.85}

\section{Implementation Details and Efficiency Protocol}
\label{app:implementation}

All RFD cells share one architecture: $K{=}3$ stages; right-aligned non-overlapping 48-step patches; an affine patch embedding with per-patch position parameters; one pre-LayerNorm Transformer block shared across stages; and a shared affine head emitting $H/K$ steps. If $L$ is not divisible by 48, the oldest remainder is discarded, not the newest suffix, so all values emitted at Stage $k$ enter Stage $k{+}1$. Operator weights are shared across channels, but recipes with positional offsets enabled also learn an input-offset tensor of shape $L\times C$; this tensor is channel-specific and explains the channel-count-dependent parameter total in Table~\ref{tab:rq4_efficiency}. Except for ETTh1, each window uses detached last-$H$ mean/std normalization; ETTh1 keeps loader-global coordinates. Training uses the complex-FFT-residual objective and prefix weights $20/k$, with Adam, StepLR, gradient clipping at 10, and checkpoint selection by full-validation MSE.

Table~\ref{tab:hyperparams} lists the frozen per-dataset recipe variables. The architecture, the stage count $K{=}3$, the weight sharing, and the objective are identical in every cell; only \texttt{pred\_len} varies across horizons. The same frozen recipes are used at every horizon. The matched-look-back comparison in Section~\ref{sec:benchmark} additionally uses a look-back-96 variant of these recipes; all locally trained results use seeds 2021/2022/2023, validation-selected checkpoints, and a separate frozen test pass. Measured parameter counts of the trained models over the $6{\times}4$ development grid range from 83,441 (Weather, $H{=}96$) to 451,057 ($H{=}720$ on ETTh1, ETTm1, and ETTm2), consistent with the 0.08--0.46M range quoted in Section~\ref{sec:approach}.

\begin{table}[h]
\centering
\caption{Frozen RFD recipe per dataset (common to all horizons). The model width $d$ equals $\min(d_{\mathrm{ff}},128)$. pe: positional offsets on the rolling window (shared across stages). $\alpha$: frequency weight of the TFMAE objective. $\gamma$: StepLR decay factor (step size 2). Budget: maximum epochs (early-stopping patience). ETTh1 uses the global-standardization variant of the model; all other datasets use anchored normalization.}
\label{tab:hyperparams}
    \begin{small}
    \resizebox{\columnwidth}{!}{%
    \begin{tabular}{lcccccccccc}
    \toprule
    \midrule
    Dataset & $L$ & $d_{\mathrm{ff}}$ ($d$) & dropout & pe & $\alpha$ & $\gamma$ & batch & lr & budget \\
    \midrule
    ETTh1   & 512 & 2048 (128) & 0.7 & on  & 0.00  & 0.3 & 128 & $6{\times}10^{-4}$ & 10 (3) \\
    ETTh2   & 512 & 128        & 0.1 & off & 0.00  & 0.3 & 16  & $5{\times}10^{-4}$ & 5 (3) \\
    ETTm1   & 512 & 2048 (128) & 0.5 & on  & 0.20  & 0.9 & 64  & $5{\times}10^{-4}$ & 5 (2) \\
    ETTm2   & 512 & 128        & 0.3 & on  & 0.45  & 0.9 & 64  & $2{\times}10^{-3}$ & 5 (2) \\
    Weather & 720 & 64         & 0.8 & on  & 0.50  & 0.1 & 64  & $4{\times}10^{-3}$ & 5 (10) \\
    Traffic & 512 & 2048 (128) & 0.5 & off & 0.20  & 0.9 & 16  & $8{\times}10^{-4}$ & 40 (15) \\
    \bottomrule
    \end{tabular}%
    }
    \end{small}
\end{table}

Table~\ref{tab:rq4_efficiency} is measured on a single NVIDIA H20 with synthetic inputs (no dataset files are read): Adam with learning rate $10^{-3}$, batch 32, input windows of length 96, and $C\in\{7,862\}$ channels. Each point runs 3 warmup iterations followed by 20 timed forward, backward, and optimizer-update iterations bracketed by \texttt{torch.cuda.synchronize}. The parameter increase from $C{=}7$ to $C{=}862$ is exactly $96(862-7)=82{,}080$, contributed by the optional $L\times C$ input-offset tensor. The forecasting operator itself shares weights across channels, but the complete model with positional offsets is not parameter-independent of $C$.

The iterative baseline column, ReNF (beta), is the official implementation (\url{github.com/Luoauoa/ReNF}, commit \texttt{c232ec5}) at a lightweight configuration ($d_{\mathrm{ff}}{=}512$, three decoder layers, one block, no dropout) chosen to keep its parameter count in the same order of magnitude as the widest RFD variant; its figures are not comparable to the official full configuration. Measured under its pinned official ETTm1 $H{=}96$ configuration ($d_{\mathrm{ff}}{=}2048$, $L{=}512$), ReNF uses 16,400,320 parameters, the source of the ${\sim}90\times$ figure in Section~\ref{sec:approach}.

\input{appendix_controlled.tex}

\section{Per-Horizon Matched Results}
\label{app:full_results}

Tables~\ref{tab:supp_full_s} and~\ref{tab:supp_full_m} provide the per-horizon breakdown of the horizon-averaged Tables~\ref{tab:baseline_avg} and~\ref{tab:ms_avg}. RFD numbers are produced by the matched-look-back ($L{=}96$) evaluation campaign over seeds 2021, 2022, and 2023, with checkpoints selected by full validation MSE and then evaluated once on the official test split. Baseline figures come from official releases or faithful re-implementations under the same look-back-96 protocol. The frozen long-context comparison in Table~\ref{tab:test_multiseed} selects its per-cell baseline only from the ten baseline families displayed in these matched tables.

\input{appendix_full_tables.tex}

\clearpage

\end{document}

%% file: avg_table_s.tex
\begin{table*}[htbp]
\centering
\caption{Univariate test MSE/MAE, averaged over four horizons at $L{=}96$.}
\label{tab:baseline_avg}
    \resizebox{\textwidth}{!}{%
    \footnotesize
    \setlength{\tabcolsep}{3.5pt}%
    \begin{tabular}{c|cc|cc|cc|cc|cc|cc|cc|cc|cc|cc|cc}
    \toprule
    \midrule
     \multirow{2}{*}{Dataset}  & \multicolumn{2}{c|}{RFD} & \multicolumn{2}{c|}{WPMixer} & \multicolumn{2}{c|}{TimeMixer} & \multicolumn{2}{c|}{PatchTST} & \multicolumn{2}{c|}{iTransformer} & \multicolumn{2}{c|}{DLinear} & \multicolumn{2}{c|}{FEDformer} & \multicolumn{2}{c|}{SparseTSF} & \multicolumn{2}{c|}{FITS} & \multicolumn{2}{c|}{TimeXer} & \multicolumn{2}{c}{TQNet} \\
     & MSE & MAE & MSE & MAE & MSE & MAE & MSE & MAE & MSE & MAE & MSE & MAE & MSE & MAE & MSE & MAE & MSE & MAE & MSE & MAE & MSE & MAE \\
    \midrule
    ETTh1 & 0.122 & 0.268 & 0.078 & 0.214 & \textbf{0.074} & \textbf{0.209} & 0.081 & 0.220 & 0.077 & 0.213 & 0.109 & 0.252 & 0.104 & 0.250 & 0.091 & 0.235 & 0.088 & 0.230 & 0.077 & 0.216 & \underline{0.076} & \underline{0.211} \\
    ETTh2 & \textbf{0.190} & \textbf{0.341} & \underline{0.192} & 0.346 & 0.202 & 0.353 & 0.201 & 0.353 & 0.195 & 0.349 & 0.219 & 0.364 & 0.204 & 0.350 & 0.260 & 0.408 & 0.235 & 0.383 & 0.204 & 0.354 & \textbf{0.190} & \underline{0.342} \\
    ETTm1 & \underline{0.054} & \textbf{0.172} & \textbf{0.053} & \underline{0.173} & \textbf{0.053} & \textbf{0.172} & \underline{0.054} & 0.174 & 0.056 & 0.177 & 0.066 & 0.188 & 0.064 & 0.196 & 0.055 & 0.179 & \textbf{0.053} & 0.174 & 0.056 & 0.176 & \textbf{0.053} & \textbf{0.172} \\
    ETTm2 & \textbf{0.120} & \textbf{0.255} & 0.122 & 0.257 & 0.123 & 0.260 & 0.123 & 0.261 & 0.133 & 0.270 & 0.125 & 0.261 & 0.128 & 0.269 & 0.136 & 0.282 & \underline{0.121} & 0.260 & 0.124 & 0.261 & \underline{0.121} & \underline{0.256} \\
    Weather & 0.647 & 0.569 & 0.657 & 0.574 & 0.648 & 0.572 & 0.656 & 0.574 & \underline{0.624} & \underline{0.558} & 0.714 & 0.590 & 0.843 & 0.704 & 0.695 & 0.605 & 0.653 & 0.580 & \textbf{0.606} & \textbf{0.551} & 0.650 & 0.571 \\
    Traffic & \underline{0.154} & \underline{0.231} & 0.178 & 0.266 & 0.246 & 0.349 & 0.181 & 0.268 & \textbf{0.142} & \textbf{0.226} & 0.300 & 0.387 & 0.184 & 0.274 & 1.168 & 0.820 & 0.677 & 0.660 & 0.156 & 0.241 & 0.178 & 0.265 \\
    \bottomrule
\end{tabular}%
    }
\end{table*}

%% file: avg_table_ms.tex
\begin{table*}[htbp]
\centering
\caption{Multivariate test MSE/MAE, averaged over four horizons at $L{=}96$.}
\label{tab:ms_avg}
    \resizebox{\textwidth}{!}{%
    \footnotesize
    \setlength{\tabcolsep}{3.5pt}%
    \begin{tabular}{c|cc|cc|cc|cc|cc|cc|cc|cc|cc|cc|cc}
    \toprule
    \midrule
     \multirow{2}{*}{Dataset}  & \multicolumn{2}{c|}{RFD} & \multicolumn{2}{c|}{WPMixer} & \multicolumn{2}{c|}{TimeMixer} & \multicolumn{2}{c|}{PatchTST} & \multicolumn{2}{c|}{iTransformer} & \multicolumn{2}{c|}{DLinear} & \multicolumn{2}{c|}{FEDformer} & \multicolumn{2}{c|}{SparseTSF} & \multicolumn{2}{c|}{FITS} & \multicolumn{2}{c|}{TimeXer} & \multicolumn{2}{c}{TQNet} \\
     & MSE & MAE & MSE & MAE & MSE & MAE & MSE & MAE & MSE & MAE & MSE & MAE & MSE & MAE & MSE & MAE & MSE & MAE & MSE & MAE & MSE & MAE \\
    \midrule
    ETTh1 & 0.455 & 0.448 & \textbf{0.442} & \textbf{0.434} & 0.463 & 0.449 & 0.464 & 0.458 & 0.463 & 0.451 & 0.457 & 0.454 & \underline{0.443} & 0.462 & 0.712 & 0.572 & 0.590 & 0.518 & 0.459 & 0.451 & 0.463 & \underline{0.443} \\
    ETTh2 & \textbf{0.371} & \textbf{0.391} & 0.388 & 0.412 & 0.378 & 0.404 & 0.392 & 0.414 & 0.393 & 0.411 & 0.556 & 0.514 & 0.443 & 0.455 & 0.442 & 0.447 & 0.419 & 0.428 & \underline{0.373} & \underline{0.400} & 0.375 & 0.401 \\
    ETTm1 & \textbf{0.382} & \textbf{0.388} & \underline{0.385} & \underline{0.395} & 0.392 & 0.402 & 0.393 & 0.407 & 0.426 & 0.419 & 0.403 & 0.407 & 0.441 & 0.455 & 0.463 & 0.438 & 0.416 & 0.409 & 0.398 & 0.407 & 0.397 & 0.402 \\
    ETTm2 & \textbf{0.274} & \textbf{0.317} & \underline{0.278} & \underline{0.323} & 0.279 & 0.326 & 0.289 & 0.335 & 0.290 & 0.333 & 0.352 & 0.400 & 0.307 & 0.352 & 0.303 & 0.341 & 0.288 & 0.329 & 0.283 & 0.327 & 0.284 & 0.327 \\
    Weather & 0.319 & \textbf{0.330} & \underline{0.316} & \underline{0.331} & 0.318 & 0.333 & 0.326 & 0.336 & 0.327 & 0.336 & 0.332 & 0.373 & 0.414 & 0.442 & 0.367 & 0.370 & 0.348 & 0.353 & \textbf{0.311} & \textbf{0.330} & 0.319 & 0.333 \\
    Traffic & 0.523 & 0.323 & 0.495 & 0.312 & 0.550 & 0.359 & \underline{0.460} & 0.292 & \textbf{0.429} & \textbf{0.284} & 0.625 & 0.385 & 0.611 & 0.374 & 0.642 & 0.381 & 0.632 & 0.382 & 0.475 & \underline{0.288} & 0.473 & 0.309 \\
    \bottomrule
\end{tabular}%
    }
\end{table*}

%% file: appendix_controlled.tex
\section{Controlled Three-Seed Analyses}
\label{app:controlled}
Every entry in this section is validation-only and reports mean $\pm$ sample standard deviation over seeds 2021/2022/2023. These controls diagnose the construction; they are not additional test-set model-selection results.

\subsection{Feedback Information and Gradient Flow}
Table~\ref{tab:supp_feedback} separates forward information from gradient flow. F0 uses ordinary feedback; F1 applies stop-gradient to the same forward values; F2 adds an explicit stage identifier while retaining feedback; and F3 keeps the stage identifier but removes predictive feedback.

\begin{table}[htbp]
\centering
\caption{Feedback controls: validation MSE (mean $\pm$ sample standard deviation).}
\label{tab:supp_feedback}
\scriptsize
\setlength{\tabcolsep}{3.8pt}
\begin{tabular}{lcccc}
\toprule
Dataset & F0: feedback & F1: detached & F2: stage+feedback & F3: stage only \\
\midrule
ETTh1 & .6699 $\pm$ .0020 & .6701 $\pm$ .0019 & .6695 $\pm$ .0019 & 1.2064 $\pm$ .0033 \\
ETTh2 & .2134 $\pm$ .0012 & .2131 $\pm$ .0012 & .2124 $\pm$ .0013 & .2513 $\pm$ .0007 \\
ETTm1 & .3812 $\pm$ .0007 & .3842 $\pm$ .0013 & .3811 $\pm$ .0009 & .3917 $\pm$ .0012 \\
ETTm2 & .1216 $\pm$ .0011 & .1217 $\pm$ .0011 & .1214 $\pm$ .0006 & .1227 $\pm$ .0002 \\
\bottomrule
\end{tabular}
\end{table}

Across all twelve paired seeds, F3 is worse than F2. The mean penalty is 80.19\% on ETTh1, 18.34\% on ETTh2, 2.79\% on ETTm1, and 1.09\% on ETTm2. By contrast, F1 differs from F0 by at most 0.79\% in mean MSE. The isolated signal is therefore the forward forecast state, not gradient propagation through that state.

\subsection{Objective Decomposition}
The four loss arms in Table~\ref{tab:supp_loss} add one design choice at a time: L0 is final-horizon MSE, L1 adds prefix supervision, L2 changes the pointwise term to MAE, and L3 adds the complex spectral residual used by RFD.

\begin{table}[htbp]
\centering
\caption{Loss controls: validation MSE (mean $\pm$ sample standard deviation).}
\label{tab:supp_loss}
\scriptsize
\setlength{\tabcolsep}{3.6pt}
\begin{tabular}{llcccc}
\toprule
Dataset & $H$ & L0: final MSE & L1: prefix MSE & L2: prefix MAE & L3: prefix TFMAE \\
\midrule
ETTh1 & 96 & .6631 $\pm$ .0007 & .6530 $\pm$ .0008 & .6699 $\pm$ .0020 & .6699 $\pm$ .0020 \\
ETTh2 & 96 & .2144 $\pm$ .0004 & .2160 $\pm$ .0024 & .2134 $\pm$ .0012 & .2134 $\pm$ .0012 \\
ETTm1 & 96 & .3918 $\pm$ .0024 & .3915 $\pm$ .0007 & .3813 $\pm$ .0012 & .3812 $\pm$ .0007 \\
ETTm1 & 720 & .9656 $\pm$ .0070 & .9544 $\pm$ .0120 & .9600 $\pm$ .0042 & .9464 $\pm$ .0018 \\
ETTm2 & 96 & .1269 $\pm$ .0005 & .1264 $\pm$ .0010 & .1234 $\pm$ .0004 & .1216 $\pm$ .0011 \\
ETTm2 & 720 & .2783 $\pm$ .0038 & .2761 $\pm$ .0012 & .2681 $\pm$ .0015 & .2629 $\pm$ .0043 \\
\bottomrule
\end{tabular}
\end{table}

The full objective is not uniformly optimal. Its clearest spectral gains occur on the 15-minute datasets, especially at $H{=}720$; the hourly recipes set the spectral weight to zero and instead expose a pointwise-metric trade-off. This supports a dataset- and horizon-dependent objective, not a universal frequency-loss claim.

\subsection{Repeated Shared Computation at Fixed Factorization}
Unlike the stage-count sweep in Section~\ref{sec:rq2}, Table~\ref{tab:supp_depth} fixes $K$, block length, head size, feedback frequency, and objective, then applies the same operator one, two, or four times inside each stage.

\begin{table}[htbp]
\centering
\caption{Fixed-factorization shared-compute control: validation MSE (mean $\pm$ sample standard deviation).}
\label{tab:supp_depth}
\scriptsize
\setlength{\tabcolsep}{4pt}
\begin{tabular}{llrccc}
\toprule
Dataset & $H$ & width & 1 repeat & 2 repeats & 4 repeats \\
\midrule
ETTh1 & 96 & 32 & .6962 $\pm$ .0020 & .6952 $\pm$ .0019 & .6935 $\pm$ .0015 \\
ETTh1 & 96 & 64 & .6816 $\pm$ .0010 & .6804 $\pm$ .0003 & .6767 $\pm$ .0026 \\
ETTh1 & 96 & 128 & .6702 $\pm$ .0028 & .6689 $\pm$ .0025 & .6675 $\pm$ .0018 \\
ETTh1 & 336 & 32 & 1.1275 $\pm$ .0011 & 1.1274 $\pm$ .0012 & 1.1278 $\pm$ .0028 \\
ETTh1 & 336 & 64 & 1.1158 $\pm$ .0050 & 1.1159 $\pm$ .0047 & 1.1150 $\pm$ .0035 \\
ETTh1 & 336 & 128 & 1.1027 $\pm$ .0156 & 1.0991 $\pm$ .0171 & 1.0959 $\pm$ .0162 \\
ETTm2 & 96 & 32 & .1229 $\pm$ .0005 & .1226 $\pm$ .0005 & .1225 $\pm$ .0007 \\
ETTm2 & 96 & 64 & .1225 $\pm$ .0006 & .1217 $\pm$ .0008 & .1222 $\pm$ .0006 \\
ETTm2 & 96 & 128 & .1216 $\pm$ .0011 & .1208 $\pm$ .0006 & .1209 $\pm$ .0008 \\
Traffic & 336 & 32 & .3458 $\pm$ .0004 & .3453 $\pm$ .0004 & .3453 $\pm$ .0003 \\
Traffic & 336 & 64 & .3415 $\pm$ .0005 & .3412 $\pm$ .0005 & .3410 $\pm$ .0015 \\
Traffic & 336 & 128 & .3393 $\pm$ .0005 & .3389 $\pm$ .0007 & .3380 $\pm$ .0006 \\
\bottomrule
\end{tabular}
\end{table}

The best repeated setting improves all 12 cells relative to one application, but only by 0.01--0.72\%. Width produces the larger movement. Repeated computation is consequently a real but secondary capacity axis in these controls.

\subsection{Operator Generality within the Same Rollout}
Table~\ref{tab:supp_operator} holds the block-rollout protocol fixed and replaces only the block operator. PT denotes the patch Transformer; ``gated'' is the identity-gated version used by RFD.

\begin{table}[htbp]
\centering
\caption{Matched block-rollout operators: validation MSE (mean $\pm$ sample standard deviation).}
\label{tab:supp_operator}
\scriptsize
\setlength{\tabcolsep}{3.5pt}
\begin{tabular}{llcccc}
\toprule
Dataset & $H$ & gated PT & ungated PT & patch linear & rolling MLP \\
\midrule
ETTm1 & 96 & .3551 $\pm$ .0017 & .3544 $\pm$ .0030 & .3763 $\pm$ .0016 & .3832 $\pm$ .0020 \\
ETTm1 & 336 & .6156 $\pm$ .0024 & .6171 $\pm$ .0054 & .6504 $\pm$ .0012 & .6468 $\pm$ .0044 \\
ETTm2 & 96 & .1148 $\pm$ .0008 & .1156 $\pm$ .0007 & .1135 $\pm$ .0005 & .1129 $\pm$ .0010 \\
ETTm2 & 336 & .2009 $\pm$ .0016 & .2020 $\pm$ .0024 & .1962 $\pm$ .0010 & .1965 $\pm$ .0037 \\
Traffic & 96 & .3197 $\pm$ .0006 & .3199 $\pm$ .0003 & .3383 $\pm$ .0004 & .3416 $\pm$ .0004 \\
Traffic & 336 & .3393 $\pm$ .0005 & .3376 $\pm$ .0004 & .3504 $\pm$ .0001 & .3934 $\pm$ .0005 \\
\bottomrule
\end{tabular}
\end{table}

A Transformer variant leads four of six cells, whereas the linear or MLP operator leads both ETTm2 cells. The rollout principle therefore transfers across operator families, while the gated Transformer is a strong default but not a universally necessary component.

%% file: appendix_full_tables.tex

\begin{table*}[htbp]
\centering
\caption{Comprehensive per-horizon test results (univariate S mode, look-back 96; lower is better). RFD is the three-seed mean (2021/2022/2023), each seed evaluated once on the official test split with the validation-selected checkpoint; baseline figures come from official releases or faithful re-implementations. Best in bold, second best underlined.}
\label{tab:supp_full_s}
\renewcommand{\arraystretch}{1.1}
\resizebox{\textwidth}{!}{%
\begin{tabular}{c|c|cc|cc|cc|cc|cc|cc|cc|cc|cc|cc|cc}
\toprule
\multirow{2}{*}{Dataset} & \multirow{2}{*}{pred\_len} & \multicolumn{2}{c|}{\textbf{RFD}} & \multicolumn{2}{c|}{\textbf{WPMixer}} & \multicolumn{2}{c|}{\textbf{TimeMixer}} & \multicolumn{2}{c|}{\textbf{PatchTST}} & \multicolumn{2}{c|}{\textbf{iTransformer}} & \multicolumn{2}{c|}{\textbf{DLinear}} & \multicolumn{2}{c|}{\textbf{FEDformer}} & \multicolumn{2}{c|}{\textbf{SparseTSF}} & \multicolumn{2}{c|}{\textbf{FITS}} & \multicolumn{2}{c|}{\textbf{TimeXer}} & \multicolumn{2}{c}{\textbf{TQNet}} \\
\cline{3-24}
 &  & MSE & MAE & MSE & MAE & MSE & MAE & MSE & MAE & MSE & MAE & MSE & MAE & MSE & MAE & MSE & MAE & MSE & MAE & MSE & MAE & MSE & MAE \\
\midrule
\multirow{4}{*}{ETTh1} & 96 & 0.073 & 0.199 & \underline{0.056} & \underline{0.179} & 0.056 & 0.180 & 0.057 & 0.181 & 0.061 & 0.189 & 0.062 & 0.184 & 0.080 & 0.218 & 0.068 & 0.202 & 0.060 & 0.188 & 0.059 & 0.185 & \textbf{0.055} & \textbf{0.178} \\
 & 192 & 0.092 & 0.227 & 0.073 & \underline{0.205} & \underline{0.073} & 0.206 & 0.078 & 0.216 & 0.074 & 0.206 & 0.086 & 0.219 & 0.101 & 0.241 & 0.084 & 0.226 & 0.080 & 0.219 & 0.079 & 0.217 & \textbf{0.072} & \textbf{0.203} \\
 & 336 & 0.113 & 0.261 & 0.087 & 0.228 & \textbf{0.082} & \textbf{0.222} & 0.088 & 0.234 & \underline{0.084} & \underline{0.223} & 0.105 & 0.251 & 0.113 & 0.264 & 0.099 & 0.250 & 0.097 & 0.247 & 0.091 & 0.238 & 0.085 & 0.227 \\
 & 720 & 0.210 & 0.384 & 0.097 & 0.244 & \underline{0.083} & \underline{0.229} & 0.101 & 0.250 & 0.088 & 0.235 & 0.184 & 0.353 & 0.123 & 0.275 & 0.111 & 0.263 & 0.116 & 0.269 & \textbf{0.080} & \textbf{0.225} & 0.090 & 0.235 \\
\midrule
\multirow{4}{*}{ETTh2} & 96 & 0.131 & \underline{0.277} & 0.136 & 0.284 & 0.136 & 0.286 & 0.144 & 0.294 & 0.140 & 0.292 & 0.131 & 0.277 & \textbf{0.130} & \textbf{0.273} & 0.213 & 0.364 & 0.148 & 0.299 & 0.139 & 0.288 & \underline{0.131} & 0.278 \\
 & 192 & \underline{0.180} & 0.331 & 0.183 & 0.337 & 0.186 & 0.340 & 0.188 & 0.342 & 0.185 & 0.341 & 0.185 & \underline{0.331} & \textbf{0.174} & \textbf{0.325} & 0.231 & 0.382 & 0.208 & 0.360 & 0.198 & 0.351 & 0.181 & 0.333 \\
 & 336 & 0.217 & 0.372 & 0.220 & 0.375 & 0.219 & 0.374 & 0.232 & 0.384 & \textbf{0.213} & \textbf{0.369} & 0.233 & 0.380 & 0.232 & 0.376 & 0.285 & 0.432 & 0.260 & 0.412 & 0.234 & 0.384 & \underline{0.217} & \underline{0.371} \\
 & 720 & 0.231 & \textbf{0.385} & \underline{0.231} & 0.387 & 0.268 & 0.413 & 0.240 & 0.392 & 0.242 & 0.394 & 0.327 & 0.468 & 0.281 & 0.425 & 0.312 & 0.453 & 0.325 & 0.463 & 0.245 & 0.393 & \textbf{0.230} & \underline{0.385} \\
\midrule
\multirow{4}{*}{ETTm1} & 96 & 0.029 & 0.127 & 0.029 & 0.127 & \textbf{0.028} & \textbf{0.125} & 0.030 & 0.130 & 0.030 & 0.130 & 0.033 & 0.134 & 0.041 & 0.157 & 0.034 & 0.140 & 0.030 & 0.130 & \underline{0.029} & \underline{0.126} & 0.029 & 0.127 \\
 & 192 & 0.045 & 0.160 & 0.044 & 0.160 & \textbf{0.044} & \textbf{0.158} & 0.046 & 0.161 & 0.046 & 0.163 & 0.056 & 0.176 & 0.054 & 0.179 & 0.047 & 0.167 & 0.044 & 0.161 & 0.045 & 0.162 & \underline{0.044} & \underline{0.159} \\
 & 336 & 0.059 & 0.185 & 0.059 & 0.186 & 0.059 & 0.186 & \underline{0.058} & \underline{0.185} & 0.067 & 0.197 & 0.076 & 0.206 & 0.071 & 0.212 & 0.059 & 0.190 & \textbf{0.057} & 0.185 & 0.059 & 0.187 & 0.058 & \textbf{0.185} \\
 & 720 & 0.081 & \textbf{0.217} & \underline{0.081} & 0.219 & 0.082 & 0.219 & 0.082 & 0.221 & 0.082 & 0.220 & 0.100 & 0.237 & 0.091 & 0.236 & 0.081 & 0.221 & \textbf{0.080} & \underline{0.218} & 0.089 & 0.228 & 0.081 & 0.218 \\
\midrule
\multirow{4}{*}{ETTm2} & 96 & 0.068 & 0.185 & 0.067 & \underline{0.183} & 0.068 & 0.188 & 0.071 & 0.194 & 0.074 & 0.195 & 0.071 & 0.193 & \textbf{0.063} & 0.188 & 0.090 & 0.229 & 0.073 & 0.198 & \underline{0.065} & 0.185 & 0.065 & \textbf{0.182} \\
 & 192 & 0.100 & \textbf{0.233} & 0.101 & 0.235 & 0.103 & 0.239 & \underline{0.100} & 0.236 & 0.113 & 0.249 & 0.105 & 0.239 & 0.108 & 0.247 & 0.117 & 0.262 & 0.103 & 0.239 & 0.103 & 0.236 & \textbf{0.099} & \underline{0.233} \\
 & 336 & \textbf{0.130} & \textbf{0.273} & 0.132 & \underline{0.274} & 0.136 & 0.282 & 0.134 & 0.277 & 0.153 & 0.294 & 0.135 & 0.278 & 0.158 & 0.310 & 0.144 & 0.294 & \underline{0.130} & 0.276 & 0.141 & 0.286 & 0.132 & 0.275 \\
 & 720 & \underline{0.181} & \underline{0.329} & 0.187 & 0.337 & 0.183 & 0.332 & 0.187 & 0.336 & 0.191 & 0.342 & 0.189 & 0.333 & 0.182 & 0.331 & 0.192 & 0.343 & \textbf{0.180} & \textbf{0.329} & 0.189 & 0.336 & 0.185 & 0.333 \\
\midrule
\multirow{4}{*}{Weather} & 96 & 0.497 & 0.488 & 0.499 & 0.488 & 0.503 & 0.493 & 0.477 & 0.480 & \underline{0.449} & \underline{0.462} & 0.551 & 0.514 & 0.582 & 0.562 & 0.568 & 0.543 & 0.514 & 0.506 & \textbf{0.436} & \textbf{0.457} & 0.498 & 0.487 \\
 & 192 & 0.608 & 0.551 & 0.606 & 0.549 & 0.597 & 0.548 & 0.607 & 0.551 & \underline{0.564} & \underline{0.531} & 0.653 & 0.568 & 1.037 & 0.818 & 0.642 & 0.582 & 0.595 & 0.553 & \textbf{0.544} & \textbf{0.521} & 0.595 & 0.545 \\
 & 336 & 0.641 & 0.573 & 0.659 & 0.582 & 0.642 & 0.575 & 0.647 & 0.580 & \underline{0.630} & \underline{0.570} & 0.740 & 0.603 & 0.797 & 0.677 & 0.683 & 0.603 & 0.646 & 0.583 & \textbf{0.604} & \textbf{0.559} & 0.651 & 0.579 \\
 & 720 & \underline{0.842} & \textbf{0.664} & 0.866 & 0.675 & 0.851 & 0.672 & 0.894 & 0.686 & 0.852 & 0.669 & 0.911 & 0.676 & 0.956 & 0.758 & 0.888 & 0.692 & 0.857 & 0.677 & \textbf{0.839} & \underline{0.668} & 0.854 & 0.671 \\
\midrule
\multirow{4}{*}{Traffic} & 96 & 0.162 & 0.234 & 0.184 & 0.272 & 0.230 & 0.334 & 0.175 & 0.263 & \textbf{0.137} & \textbf{0.217} & 0.328 & 0.417 & 0.170 & 0.260 & 1.357 & 0.926 & 0.438 & 0.525 & \underline{0.150} & \underline{0.233} & 0.180 & 0.267 \\
 & 192 & \underline{0.151} & \underline{0.226} & 0.174 & 0.259 & 0.244 & 0.349 & 0.174 & 0.262 & \textbf{0.139} & \textbf{0.218} & 0.279 & 0.368 & 0.181 & 0.265 & 0.842 & 0.661 & 0.627 & 0.639 & 0.152 & 0.237 & 0.173 & 0.257 \\
 & 336 & \underline{0.145} & \textbf{0.224} & 0.168 & 0.257 & 0.245 & 0.347 & 0.173 & 0.263 & \textbf{0.138} & \underline{0.225} & 0.274 & 0.365 & 0.170 & 0.265 & 1.355 & 0.900 & 0.723 & 0.698 & 0.152 & 0.239 & 0.169 & 0.258 \\
 & 720 & \underline{0.160} & \textbf{0.239} & 0.188 & 0.276 & 0.266 & 0.365 & 0.201 & 0.285 & \textbf{0.154} & \underline{0.246} & 0.319 & 0.399 & 0.215 & 0.306 & 1.120 & 0.795 & 0.919 & 0.777 & 0.169 & 0.255 & 0.188 & 0.277 \\
\bottomrule
\end{tabular}%
}
\end{table*}

\begin{table*}[htbp]
\centering
\caption{Comprehensive per-horizon test results (multivariate-to-multivariate M mode, look-back 96; lower is better). RFD is the three-seed mean (2021/2022/2023), each seed evaluated once on the official test split with the validation-selected checkpoint; baseline figures come from official releases or faithful re-implementations. Best in bold, second best underlined.}
\label{tab:supp_full_m}
\renewcommand{\arraystretch}{1.1}
\resizebox{\textwidth}{!}{%
\begin{tabular}{c|c|cc|cc|cc|cc|cc|cc|cc|cc|cc|cc|cc}
\toprule
\multirow{2}{*}{Dataset} & \multirow{2}{*}{pred\_len} & \multicolumn{2}{c|}{\textbf{RFD}} & \multicolumn{2}{c|}{\textbf{WPMixer}} & \multicolumn{2}{c|}{\textbf{TimeMixer}} & \multicolumn{2}{c|}{\textbf{PatchTST}} & \multicolumn{2}{c|}{\textbf{iTransformer}} & \multicolumn{2}{c|}{\textbf{DLinear}} & \multicolumn{2}{c|}{\textbf{FEDformer}} & \multicolumn{2}{c|}{\textbf{SparseTSF}} & \multicolumn{2}{c|}{\textbf{FITS}} & \multicolumn{2}{c|}{\textbf{TimeXer}} & \multicolumn{2}{c}{\textbf{TQNet}} \\
\cline{3-24}
 &  & MSE & MAE & MSE & MAE & MSE & MAE & MSE & MAE & MSE & MAE & MSE & MAE & MSE & MAE & MSE & MAE & MSE & MAE & MSE & MAE & MSE & MAE \\
\midrule
\multirow{4}{*}{ETTh1} & 96 & 0.390 & 0.400 & \textbf{0.375} & \textbf{0.394} & 0.402 & 0.413 & 0.393 & 0.413 & 0.386 & 0.404 & 0.389 & 0.404 & \underline{0.380} & 0.422 & 0.749 & 0.577 & 0.469 & 0.453 & 0.397 & 0.412 & 0.383 & \underline{0.398} \\
 & 192 & 0.442 & 0.434 & \underline{0.428} & \textbf{0.425} & 0.452 & 0.440 & 0.447 & 0.439 & 0.443 & 0.434 & 0.438 & 0.433 & \textbf{0.421} & 0.447 & 0.659 & 0.539 & 0.531 & 0.486 & 0.434 & 0.436 & 0.444 & \underline{0.428} \\
 & 336 & 0.482 & 0.457 & \textbf{0.467} & \textbf{0.442} & 0.497 & 0.461 & 0.491 & 0.473 & 0.492 & 0.463 & 0.488 & 0.467 & \underline{0.471} & 0.475 & 0.736 & 0.583 & 0.615 & 0.530 & 0.490 & 0.461 & 0.491 & \underline{0.452} \\
 & 720 & 0.507 & 0.501 & \textbf{0.497} & \textbf{0.475} & \underline{0.500} & \underline{0.483} & 0.524 & 0.506 & 0.531 & 0.504 & 0.513 & 0.511 & 0.501 & 0.502 & 0.706 & 0.589 & 0.743 & 0.602 & 0.514 & 0.496 & 0.534 & 0.495 \\
\midrule
\multirow{4}{*}{ETTh2} & 96 & \textbf{0.286} & \textbf{0.331} & 0.298 & 0.348 & \underline{0.291} & 0.343 & 0.318 & 0.361 & 0.294 & 0.346 & 0.342 & 0.396 & 0.351 & 0.392 & 0.381 & 0.406 & 0.317 & 0.360 & 0.295 & \underline{0.342} & 0.293 & 0.343 \\
 & 192 & \textbf{0.369} & \textbf{0.383} & 0.381 & 0.402 & 0.375 & \underline{0.395} & 0.392 & 0.408 & 0.380 & 0.399 & 0.473 & 0.473 & 0.442 & 0.450 & 0.443 & 0.439 & 0.409 & 0.415 & 0.381 & 0.402 & \underline{0.373} & 0.395 \\
 & 336 & \textbf{0.413} & \textbf{0.419} & 0.424 & 0.437 & 0.419 & 0.432 & 0.419 & 0.436 & 0.434 & 0.436 & 0.584 & 0.535 & 0.499 & 0.490 & 0.477 & 0.468 & 0.456 & 0.454 & \underline{0.415} & \underline{0.426} & 0.422 & 0.431 \\
 & 720 & 0.415 & \underline{0.433} & 0.451 & 0.460 & 0.426 & 0.445 & 0.438 & 0.454 & 0.465 & 0.465 & 0.823 & 0.654 & 0.480 & 0.487 & 0.470 & 0.472 & 0.493 & 0.484 & \textbf{0.402} & \textbf{0.431} & \underline{0.411} & 0.434 \\
\midrule
\multirow{4}{*}{ETTm1} & 96 & \textbf{0.313} & \textbf{0.345} & \underline{0.317} & \underline{0.354} & 0.330 & 0.370 & 0.333 & 0.370 & 0.346 & 0.375 & 0.345 & 0.373 & 0.378 & 0.420 & 0.429 & 0.419 & 0.357 & 0.377 & 0.327 & 0.365 & 0.331 & 0.367 \\
 & 192 & \textbf{0.364} & \textbf{0.376} & \underline{0.365} & \underline{0.380} & 0.369 & 0.388 & 0.366 & 0.388 & 0.398 & 0.401 & 0.381 & 0.391 & 0.425 & 0.444 & 0.434 & 0.419 & 0.394 & 0.394 & 0.383 & 0.393 & 0.368 & 0.384 \\
 & 336 & \textbf{0.395} & \textbf{0.397} & \underline{0.398} & \underline{0.405} & 0.402 & 0.409 & 0.408 & 0.418 & 0.448 & 0.433 & 0.414 & 0.413 & 0.460 & 0.467 & 0.469 & 0.443 & 0.426 & 0.415 & 0.408 & 0.413 & 0.414 & 0.412 \\
 & 720 & \textbf{0.456} & \textbf{0.432} & \underline{0.460} & 0.442 & 0.465 & \underline{0.441} & 0.465 & 0.450 & 0.511 & 0.468 & 0.473 & 0.451 & 0.503 & 0.488 & 0.519 & 0.471 & 0.486 & 0.449 & 0.474 & 0.456 & 0.476 & 0.447 \\
\midrule
\multirow{4}{*}{ETTm2} & 96 & \textbf{0.171} & \textbf{0.250} & \underline{0.174} & \underline{0.255} & 0.176 & 0.261 & 0.181 & 0.267 & 0.185 & 0.266 & 0.192 & 0.291 & 0.201 & 0.286 & 0.204 & 0.284 & 0.185 & 0.268 & 0.176 & 0.259 & 0.177 & 0.259 \\
 & 192 & \textbf{0.235} & \textbf{0.293} & \underline{0.240} & \underline{0.299} & 0.242 & 0.304 & 0.250 & 0.311 & 0.253 & 0.309 & 0.288 & 0.364 & 0.267 & 0.329 & 0.263 & 0.318 & 0.248 & 0.306 & 0.253 & 0.308 & 0.242 & 0.301 \\
 & 336 & \textbf{0.297} & \textbf{0.333} & 0.300 & \underline{0.338} & \underline{0.298} & 0.339 & 0.310 & 0.351 & 0.312 & 0.350 & 0.383 & 0.427 & 0.335 & 0.371 & 0.323 & 0.355 & 0.309 & 0.343 & 0.302 & 0.342 & 0.307 & 0.343 \\
 & 720 & \textbf{0.393} & \textbf{0.390} & 0.400 & 0.400 & \underline{0.399} & 0.398 & 0.416 & 0.412 & 0.413 & 0.407 & 0.543 & 0.516 & 0.427 & 0.421 & 0.421 & 0.408 & 0.409 & 0.398 & 0.400 & \underline{0.398} & 0.411 & 0.404 \\
\midrule
\multirow{4}{*}{Weather} & 96 & 0.226 & 0.260 & 0.222 & 0.261 & 0.223 & 0.264 & 0.227 & 0.264 & 0.231 & 0.264 & 0.253 & 0.306 & 0.297 & 0.365 & 0.281 & 0.313 & 0.259 & 0.291 & \textbf{0.213} & \textbf{0.257} & \underline{0.220} & \underline{0.259} \\
 & 192 & 0.283 & \underline{0.308} & \underline{0.277} & \textbf{0.307} & 0.281 & 0.309 & 0.290 & 0.315 & 0.288 & 0.312 & 0.303 & 0.351 & 0.459 & 0.486 & 0.335 & 0.350 & 0.313 & 0.331 & \textbf{0.276} & 0.308 & 0.280 & 0.308 \\
 & 336 & 0.340 & \textbf{0.349} & \underline{0.339} & 0.352 & 0.342 & 0.353 & 0.352 & 0.356 & 0.352 & 0.358 & 0.349 & 0.387 & 0.414 & 0.439 & 0.387 & 0.384 & 0.370 & 0.369 & \textbf{0.332} & \underline{0.350} & 0.344 & 0.354 \\
 & 720 & 0.428 & \textbf{0.404} & 0.428 & 0.407 & 0.428 & 0.406 & 0.434 & 0.409 & 0.438 & 0.412 & \underline{0.424} & 0.447 & 0.484 & 0.477 & 0.466 & 0.431 & 0.452 & 0.420 & \textbf{0.422} & \underline{0.405} & 0.433 & 0.412 \\
\midrule
\multirow{4}{*}{Traffic} & 96 & 0.513 & 0.316 & 0.471 & 0.304 & 0.538 & 0.356 & \underline{0.437} & 0.281 & \textbf{0.397} & \textbf{0.269} & 0.650 & 0.399 & 0.587 & 0.356 & 0.671 & 0.396 & 0.656 & 0.395 & 0.446 & \underline{0.273} & 0.442 & 0.293 \\
 & 192 & 0.516 & 0.323 & 0.481 & 0.305 & 0.542 & 0.357 & \underline{0.446} & 0.285 & \textbf{0.419} & \textbf{0.278} & 0.599 & 0.372 & 0.608 & 0.381 & 0.616 & 0.368 & 0.607 & 0.370 & 0.468 & \underline{0.284} & 0.455 & 0.298 \\
 & 336 & 0.517 & 0.320 & 0.496 & 0.310 & 0.542 & 0.356 & \underline{0.461} & 0.291 & \textbf{0.435} & \textbf{0.285} & 0.606 & 0.375 & 0.608 & 0.367 & 0.622 & 0.370 & 0.614 & 0.372 & 0.480 & \underline{0.290} & 0.477 & 0.312 \\
 & 720 & 0.544 & 0.334 & 0.530 & 0.329 & 0.579 & 0.369 & \underline{0.496} & 0.312 & \textbf{0.466} & \textbf{0.303} & 0.646 & 0.396 & 0.640 & 0.390 & 0.660 & 0.389 & 0.651 & 0.390 & 0.506 & \underline{0.307} & 0.518 & 0.334 \\
\bottomrule
\end{tabular}%
}
\end{table*}